\documentclass[runningheads]{llncs}
\usepackage{graphicx}
\usepackage{listings}
\usepackage{array}
\begin{document}
\title{Emergent social NPC interactions in the Social NPCs Skyrim mod and beyond}
\author{Manuel Guimar\~{a}es \inst{1}\orcidID{0002-4538-1009} \and
Pedro A. Santos\inst{1,2}\orcidID{0002-1369-0085} \and
Arnav Jhala\inst{3}\orcidID{2222--3333-4444-5555}}
\authorrunning{Guimaraes et al.}
\institute{INESC-ID, Lisbon, Portugal \\
\url{https://tecnico.ulisboa.pt/pt/}
\and Instituto Superior T\'ecnico, 
Universidade de Lisboa, 
Lisbon, Portugal\\ 
\url{https://www.inesc-id.pt/} \and
Department of Computer Science, 
Visual Narrative Cluster, 
North Carolina State University, 
Raleigh, North Carolina, USA
\url{https://facultyclusters.ncsu.edu/clusters/visual-narrative/}}

\maketitle              % typeset the header of the contribution
\begin{abstract}
\textbf{Disclaimer: The paper presented here is part of discontinued issue of Game AI Pro 4} 
This work presents an implementation of a social architecture model for authoring Non-Player Character (NPC) in open world games inspired in academic research on agent-based modeling. Believable NPC authoring is burdensome in terms of rich dialogue and responsive behaviors. 

We briefly present the characteristics and advantages of using a social agent architecture for this task and describe an implementation of a social agent architecture – CiF-CK– released as a mod Social NPCs \footnote{https://steamcommunity.com/sharedfiles/filedetails/?id=751622677} for The Elder Scrolls V: Skyrim. 

\keywords{Authoring tools for Agent Modeling, NPC Agents in Games, Social Architecture Model, Social Modeling in Agents}
\end{abstract}
\section{Introduction}

Socially aware NPCs need to at least maintain their emotional status in response to observed events and their reactions based on their personality, social norms, and relationships with participants involved in the events. The credibility and believability of NPCs requires characters to express basic human traits like emotions and the ability to take actions that are responsive to the context \cite{loyall1997believable}. One of the important human traits is our social ability and awareness. People's thoughts, feelings, and behaviors are influenced by the presence, even if just imagined, of other human beings. This makes rich agent modeling in terms of social presence important for human interaction within these environments. One key aspect that has been discussed in game design literature on agent models is affinity with the player's social concerns and behaviors \cite{isbister2006better}.

Social agent architectures/models that lead to emergent NPC interactions open up rich narrative design spaces for players to explore. However, current academic architectures of this nature are embedded in well-curated designed scenarios that are not easy to generalize and scale up to commercial games. The models that are deployed successfully within games, allow the system to automatically manage and keep up with the complexity of social interactions, reducing the number of experiences that need to be explicitly authored \cite{McCoy:2009oq}. Reasoning about the social context in terms of relationship goals and desires, social status, and emotional changes is central to believable behavior \cite{goffman2005interaction}. The purpose of a social intelligence architecture is to lighten the burden on the designers and developers. Instead of scripting all interactions with each NPC, an architecture with parameterized psychological and personality characteristics of individuals, allows tuning of appraisal rules for observed in-game events, defines means of providing a variety of responses in reaction to the appraised states, and allows updates to internal states and social affinity toward other characters (including players). Given the three main inputs as described above and the reasoning mechanisms, it is possible to create deep, rich and emergent experiences for the player by focusing the authoring effort on Traits and Appraisal Rules.

\section{CIF-CK}
\begin{figure}
\includegraphics[width=\textwidth]{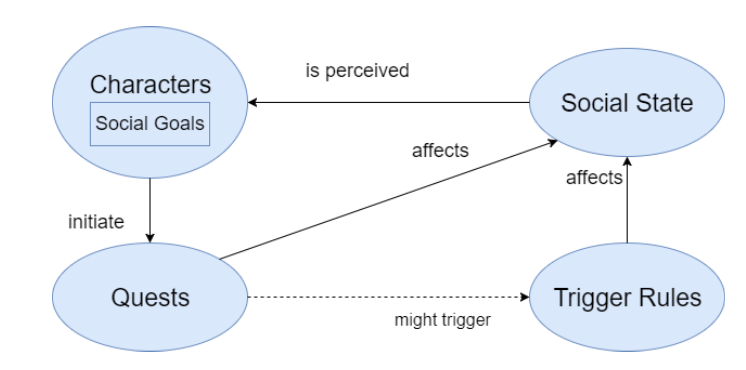}
\caption{The CiF-CK architecture} \label{img:cif}
\end{figure}

he CIF-CK (Comme il Faut Creation Kit) architecture was created to allow the application of the CIF architecture pioneered in the game Prom Week \cite{McCoy2013PromWD} to roleplaying games \footnote{We would like to have it called CIF-RPG; in fact that name had already been used for another variant and to avoid confusion we baptized our variant CIF-CK.}. The Comme il Faut architecture is described in detail in \cite{mateas2013architecture} and has four essential components (depicted in Figure \ref{img:cif}): Characters, the Social State, Social Exchanges and finally the Trigger Rules. We will now describe how each essential component was adapted along with some innovations to CiF made in CiF-CK .

\begin{table}
\caption{CIF to CIF-CK translation}\label{tab1}
\begin{tabular}{ | m{5em} | m{4.5cm}| m{6cm} | } 
\hline
CIF &  Function & CIF-CK\\
\hline
Name &  Unique Identifier & Unique Identifier \\ \hline
Gender & Agent’s Gender & Used once the Quest has finished \\ \hline
Traits &  Permanent Traits &  Permanent values that affect the decision making \\ \hline
Status &  Temporary Traits & Temporary values that affect the decision making \\ \hline
Prospective Memory & Set of desires of social exchanges with specific goals & Set of quests with specific actors
(Targets) and volitions 
 \\
\hline
\end{tabular}
\end{table}

Each character has Traits and Status that influence his/her decision making. The Traits of a character are permanent characteristics or properties that impact social exchanges such as being ``stubborn'' or ``shy''. Traits directly affect the agent’s goals and their decision making, for instance if someone is shy it will have lower social ambitions. The Status of a character captures temporary properties, e.g.: being ``drunk'', or the agent’s mood that is the result of a social exchange, such as being happy or being angry at someone.

\subsubsection{Social State}
In CIF, the social state of the world is captured by four different representations: Social Networks, Relationships, the Cultural Knowledge Base, and the Social Facts Knowledge
Base \cite{mateas2013architecture}. Because of the large dimension of these types of games, RPGs, its Game Engines render and compute only what is in the surrounding area of the Player \cite{rabin2006ai}. Any script associated with an NPC that is in a different place as the Player will not be processed. As a consequence in any implementation NPCs must consider and be considered if they are in the same area as the Player. In order to store all the data the Social State captures one should use a Static entity that can be accessed by anyone at any time. While in CiF the Social State is public, in CiF-CK the Social Networks and preferences of each character are private and should not be accessed by a different character.

In CiF-CK we added a simple Theory of Mind feature to the Comme il Faut architecture, with believes about what the state of other characters Social Networks. We considered that the Social State is something Characters perceive, however, it might not be the actual reality
\cite{haddadi1996belief}. For example, Sarah might think John likes her, despite the fact that John actually hates her. Because Sarah believes John likes her, she will act accordingly.

\subsubsection{Social Exchanges}
All Social Exchanges are the possible actions that the characters perform and have different preconditions and consequences. It is up to the Characters to decide what they want to do. In Comme il Faut each Social Exchange has a value called volition. In most Role Playing Games everything revolves around quests. Usually quests have pre-conditions to start, stages, conditions to end (successfully or not), and effects. On the other hand,  the primary knowledge representation element in CiF is the Social Exchange, a collection of patterns of (primarily dialogue) interaction where the exact performance and social outcome varies based on the personality-specific attributes of the characters involved and the current social state \cite{mateas2013architecture}.

The similarity between Quests, in RPGs, and Social Exchanges, in CiF, allow us to adapt one to the other. We can use Quests the same way Social Exchanges are used in CiF, with some adaptations. We exhibit this transition in Table 2. If the game engine one is using has a general implementation of quests, where those can be also started by the NPCs (as Bethesda RPGs have) then those can be used directly. In a game with a less general implementation of quests, some extension might be needed.

\begin{table}
\caption{CIF to CIF-CK translation}\label{tab1}
\begin{tabular}{ | m{5em} | m{4.5cm}| m{6cm} | } 
\hline
CIF &  Function & CIF-CK\\
\hline
Name &  Unique Identifier & Unique Identifier \\ \hline
Gender & Agent’s Gender & Used once the Quest has finished \\ \hline
Traits &  Permanent Traits &  Permanent values that affect the decision making \\ \hline
Status &  Temporary Traits & Temporary values that affect the decision making \\ \hline
Prospective Memory & Set of desires of social exchanges with specific goals & Set of quests with specific actors
(Targets) and volitions 
 \\
\hline
\end{tabular}
\end{table}

\subsubsection{Trigger Rules}
In CiF, Trigger Rules can be ``fired'' at any point in the game and have cascading effects in the Social State  \cite{mateas2013architecture}. In order to be more efficient and avoid stressing the Game Engine, CiF-CK only verifies the trigger rules when a Social Exchange ends. In this model, quests are the only component that directly influences the social state, as such, CiF-CK runs all trigger rules when a quest ends, either by failing or succeeding, making sure if one of them triggers we can apply it and its consequences to the Social State right afterwards.

To illustrate how the architecture works with an example,  imagine that we want the game’s NPCs to initiate (or respond to) a flirt with other character (either the player or another NPC) depending on their Extrovert trait and how much they feel attracted to that character. attractiveness can be dynamic (using traits, e.g.: Likes(Female), Likes(Human), Dislikes(Blonde) ) or given directly via a social network. A Social exchange Flirt should then be created and be made available to the characters (as a possible quest). A numeric Volition then must be calculated. A simple rule can be:

\begin{lstlisting}
Volition(Flirt, x, y) 
 v=attracted_to (x,y) ; base value given by the social network
  For each trait do
	   if x.likes(trait) and y.has(trait) then
		  v += 1
	   else if x.dislikes(trait) and y.has(trait) then
		  v -= 2
		end
	end
	if v > 0 and x.has(Extrovert) then v += 2
	return v
end


\end{lstlisting}

In fact, in CiF, the volition of each Social Exchange is the sum of all active starting conditions and the above code can be abstracted.  Each starting condition is called a Microtheory. The general structure of each volition calculation function is:

\begin{lstlisting}
SUM  = 0
if Microtheory 1 is true then
SUM += Value
if Microtheory 2  is true then
SUM += Value
...
\end{lstlisting}

All the other Social Move volitions are calculated in a similar structure. In the decision making process a character calculates the volition of the possible actions and chooses to execute the one with the highest positive volition. The rule on the extrovert could be abstracted into a Microtheory with a rule that extrovert characters would add volition to certain types of social exchanges. If a character would initiate a Flirt social exchange with the player, a dialogue action would appear for the player to respond. In the case of an NPC, it could use the same volition calculus to determine either a positive or negative answer… In any case, depending on the response different Social State changes might be applied. These include improving the NPCs’ relationships and perspective of each other. Finally, some Trigger Rules might be fired. These would bring even more changes to the Social State. In our example, if the Flirt Social Move succeeded, x attraction for y increases, in turn, x beliefs that y likes her (in case its response was positive) also increases.

\section{The \textit{Social NPCs} Mod}
In this section we present a CiF-CK application example as a mod for the popular video game: ``The Elder Scrolls V: Skyrim'' \cite{skyrim} and describe its implementation process.

\subsection{The Platform: The Elder Scrolls V: Skyrim }
The Elder Scrolls V: Skyrim is an action role-playing open world video game developed by Bethesda Game Studios and published by Bethesda Softworks. It is the fifth installment in The Elder Scrolls series, it was released worldwide on November 11, 2011, and received critical acclaim. It has sold over 30 million units and years after its launch it still has a broad and active user base, partially thanks to the ``remastered editions'' and very popular the mod community. The game was developed using the Creation Engine, that was specifically rebuilt for the game. Bethesda decided to release a modding tool called Creation Kit to the gaming community. 

The release of Creation Kit to the public led to the generation of a lot of user created content There are more than 40 000 mods created by the community, available for anyone to download online for free. Mods range from smaller ones that fix bugs or tweak maps to larger ones that introduce new islands with quests or make significant UI changes.

\subsection{Creation Kit}
Creation Kit development tool is a powerful piece of software as it is the same tool that Bethesda itself used to create Skyrim. Almost everything the developers used is available
to the ``modder''. Users can use those resources or add new ones such as new textures and models. Plugins, or ``.esp'' files, are smaller collections of data which can be loaded ``on top'' of master files. These type of roles allow us to implement our architecture on top of the already built game world without overwriting the original. There are some practical concerns with modding through a scripting language. Papyrus is the scripting language for CK modding. It is quite powerful in general but mod authors are restricted to using the data structures that are provided in the scripting language. Also, Papyrus in particular can be slow and inefficient in executing scripts that have complex data structures and calls.

\subsubsection{AI Packages}
In order to give Non-Playable Characters it's behavior and not have to manage it constantly, the Creation Engine gives every actor a list of AI Packages that the Actor will run under normal circumstances. Each Package represents a behavior that the Actor will perform under certain conditions. All Actors have a Package Stack, an ordered list of potential behaviors. Periodically, the game will reevaluate each Actor’s Package Stack. The topmost package in the stack whose conditions are satisfied will be run.
This is how most NPCs return to their homes when the day is over in Skyrim. The NPC evaluates his Package Stack checking the conditions on each Package from the top down. If the conditions are satisfied in the first one, let’s say it says NPC should go home if it is past midnight then that package proceeds to carry out the proper behavior.

\subsubsection{Quests}
Quests are the stories and adventures of Skyrim. The original Skyrim game has hundreds and hundred of quests. Quests have multiple functions besides delivering a narrative. Some exist to simply store dialogue, others to manage random events that happen in the world. There are other quests define complex and important storylines and that also create secondary quests to deal with its consequence in the world. The list of quests appear in the Object Window under Character. The main components of a quest are:
\begin{itemize}
    \item Quest ID, identifies the quest and is unique to the whole game
    item Quest Stages, associated with an integer number, represent the phases of a quest, all of the stages can have starting conditions and ending conditions, for every quest there is always a starting stage and an ending stage
    \item Quest Alias, are the references to the objects that are used during the quest, they might be characters, items and even other quests. 
    \item Quest Scripts are used to apply the effects of the quest in the Game World, often used to move from stage to stage and to execute scenes.
    \item Quest Scenes, the physical performance of a quest, if a quest needs it. Scenes use temporary AI packages that override the others. For example, making a farmer deliver a letter to the Player.
\end{itemize}

Quest Alias are names or tags assigned to actors, objects, and locations used by the quest. This allows various data elements (script, packages, dialogue) to be tagged to the alias rather than to a specific object in the world, allowing quests at runtime to dynamically specify their aliases instead of being predefined. This particular ability makes it possible to reuse each Quest but with different participants/values.

\subsection{Implementing CiF-CK in Skyrim}
Our goal was to implement the CiF-CK architecture and release the resulting product online for players to test and experiment with. This section will focus on describing the mod implementation \cite{guimaraes2017prom}.

In the resulting mod we created a Game Manager in order to better handle the information each Character and each Social Move/ Quest sends and receives. We used that entity to also manage the Social State and to store information that anyone can access. The Creation Engine, that powers Skyrim, pushed the limits of hardware at the time and was designed with optimizations for the time that the game was released in 2011. As such, even on modern hardware the engine requires management of social state only within the location of the player. Consequently, it makes sense that the GameManager entity is attached to the Player’s own entity in the world. 

In Creation Kit the Player is represented in the world as an Actor with the name ``Player''. It is just like any other NPC but it’s actually controllable. The Game Manager will be represented by a script that is attached to that NPC.
The Game Manager has several functions:
\begin{itemize}
    \item Manages the Social State by storing and updating the actual Social Networks’ values and keeping track of each others public relationships.
    \item Manages the Social Exchanges’ queue, deciding which NPC is going to act next according to the queue.
    \item After a Social Exchange has finished it notifies all other NPCs of what has happened.
\end{itemize}

As we mentioned before a Social Exchange directly translates into a Quest. We call these ``Social Move'' Quests. In the resulting mod we ended up with 12 different ``Social Move '' Quests. They all have the same structure and follow the same logic. Each quest has at least 6 different stages:

\begin{table}
\caption{Social Move Quest Stages}\label{tab1}
\begin{tabular}{ | m{5em} | m{4.5cm}| m{6cm} | } 
\hline
Stage &  Description & Function\\
\hline
0 &  Normal State of a Quest (dormant) &
Waiting for the start signal \\ \hline
1 & Quest is initiated by the GameManager & Updates its Alias with references to actual NPCs (an Initiator and a Target) \\ \hline
2 & Initiate the Social Exchange Performance & Calculates the result of the exchange Starts the proper quest scene \\ \hline
3 & Social Move was successful & Inform the GM of the result \\ \hline
4 & Social Move was unsuccessful& Inform the GM of the result \\ \hline
-1 & Quest failed due to an unforeseen error & Reset itself
 \\
\hline
\end{tabular}
\end{table}

\subsubsection{Social Move Quest Scenes }
As we mentioned all Social Move Quests have at least two scenes, a successful and a failure scene. Additionally, every scene has at least three different phases:
``Go-To'' Phase: the Initiator moves near the Target
``Performance'' Phase: the Initiator performs (most often as a conversation) the interaction to the Target.
``Response'' Phase: the Target responds to the performance(might be by fighting, running away or, most often than not, a simple dialogue line).
The performance and response are generally done through dialogue however there are some exceptions, in Social Moves such as Offering a Gift and Fighting.  Each Social Move Quest has different scenes, according to the possible responses to that scene. There at least three different scenes for each Social Exchange, a failure, a success and a scene for when the Player is the target of the Exchange, where have the ability to choose his/her response.  Ideally the result of the exchange would be computed after the Initiator performed his action, for example, used a pick up line on the Target. However this puts too much of a strain on the engine and the NPC would take too much time to respond. As such, the Social Move Quests compute the result of the exchange first and only afterwards do they launch the Social Exchange Scene. 

\subsubsection{Characters}
Characters in Skyrim already have a somewhat emergent behavior. In order to transform the Characters into Social ones we had to add two scripts to each and some keywords. The main script is used for manage a character’s goals, beliefs, likes and dislikes and general information driven methods related to the Game Manager. The second script’s function is to compute the volition for each Social Exchange according to the first one’s request. Both of them allow us to properly add the behavior we need characters to have, as we described in Section 2. By adding these scripts to currently existing characters in Skyrim we are able to generate more expressive emergent social behavior. 

\subsubsection{Beliefs and Social Networks}
In CIF, Social Networks are scalar, non-reciprocal feelings characters feel towards others. Our model supports any number of Social Networks. In our implementation however, we
used only two: Attraction and Friendship. These networks model the relation of social attraction and friendship, first studied by Moreno \cite{moreno1934shall}, which reflects the affective ties that one person establishes with the others.

\subsubsection{Traits and Status}
Traits can affect both the Social Goals of an NPC and the volition of a Social Exchange. We implemented them as keywords that can be attached to Characters in Creation Kit. This allows the author to easily change an NPC’s traits by changing the keywords attached to him/her. In turn, this will change their personality and consequently the social dynamics of the location he/she is in.
 In the resulting mod we implemented five different Traits such as ``Friendly and Hostile and four different types of Status such as, ``Angry At'' and ``Drunk'', each one influences the NPCs decision making differently  \cite{guimaraes2017cif}.

\subsubsection{Forming Goals}
When the Player arrives or enter a new location the Social NPCs around him/her form their Social Goals. In our implementation, the Social Goals are influenced by the Race, Gender, Sexual Orientation and Traits of themselves and the others surrounding them. For example to determine an NPC’s attractiveness level we created an attraction Trait with the keyword,  ``cif\_t\_attractive'', that can be attached to Characters in Creation Kit. When NPCs are creating their Social Goals they will be more likely to have higher Romantic goals towards the Attractive characters (and the reverse for ugly ones).

\subsection{Gameplay with Emergent Interactions}
Imagine the player enters a new location, a house, where there are two different Social NPCs, Sabjorn and Ysolda. When the location is loaded for the first time the NPCs calculate their Social Goals. Sabjorn has the ``Friendly'' Trait, therefore he will want to be friends with everyone, including the player. Ysolda has the ``Attractive'' keyword, therefore Sabjorn will be attracted to her and have a high Romantic Goal, however, Sabjorn is from a different race, therefore Ysolda will not reciprocate Sabjorn’s feelings. 

Once the Goals have been computed it is now time to decide what everyone wants to do. NPCs will compute their volitions for every Social Exchange, pick the highest one, inform the Game Manager and then wait for further instructions (continuing their normal Skyrim functions). In this case Sabjorn will probably want to perform a ``Flirt'' towards Ysolda. Sabjorn’s scripts will tell the Game Manager to queue the Social Exchange Flirt, with him as a Initiator and Ysolda as the target. 

If there are no other exchanges in the queue, the Game Manager will start the Social Move Flirt Quest with its respective initiator and target. The quest will immediately calculate the result of the exchange and start a Quest Scene corresponding to that result. In this case the scene will be Sabjorn moving towards Ysolda, and saying: ``Your presence lights this room''. Because she has no romantic attraction towards Sabjorn her response will be neutral: ``You are too kind'', she will not immediately reject Sabjorn’s feelings but lead him on for a bit. If Sabjorn performs that action a couple of times she will start to completely reject him. After the Social Exchange is completed the Game Manager will be informed of its result and inform its participants of the result.

The player can interact with the NPCs anytime he/she wants and perform all the Social Exchanges the NPCs can. Additionally NPCs will also perform Social Moves towards the player and the he/she can decide what response they will give. At any time, the player can step in to either help Sabjorn by telling Ysolda that he is a good man or to ``ruin'' him even further by bad-mouthing him.

\subsection{Play Scenarios}
The CiF-CK architecture we implemented is used by seven different NPCs of two different locations within the game. Each one plays slightly different from the other:
Quest Scenario: the first scenario is a small Narrative experience that players can work around, using CiF-CK, with specific Characters and within a very specific drama. The idea behind this scenario to give the players a taste of what some of stories and Quests could be if they used a similar Narrative Structure to normal Skyrim Side-Quest but with the ``Social NPCs'' instead. 
Open Scenario: the second scenario is a place where the player can experience the addition of Social Ability to already existing NPCs within a more open and ``sand-box'' situation. The idea behind this is to try to understand if players interact more than they used to with previous NPCs. It is also of our interest to understand if players, without any special motivation, can create storylines on their own.

\subsection{Social NPCs Release and Reception}
The resulting mod was released online and it featured 12 different Social Moves. From ``Flirt''and ``Compliment'' to ``Insult'' and ``Fight'' and even others such as ``Spread Rumours'' and ``Offer Gifts''. Players could also perform any of these actions as seen in figure 3. Additionally we implemented 5 different Traits such as, ``Friendly'' and ``Hostile'', and 4 different types of Status such as, ``Angry At'' and ``Drunk'', each one influences the NPCs decision making and social goals formation. We can easily change an NPC's traits by changing the keywords attached to them which in turn will change their personality and consequently its decision making.

\begin{figure}
\includegraphics[width=\textwidth]{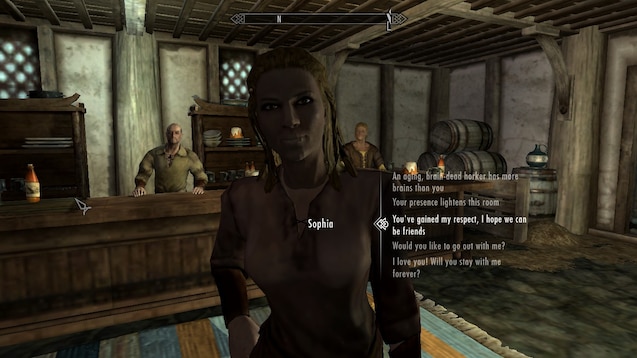}
\caption{Social Npcs Mod Gameplay} \label{img:social}
\end{figure}

\subsection{Mod Release and Results}
The mod containing CiF-CK implementation and the player scenarios is called ``Social NPCs'' and was released in the 26th of August (2016), both in Steam Workshop and in the popular mod website: ``Nexus Mods''.

It's release was quickly met with enthusiasm and positive feedback. In its first day it reached the status of ``Top Mod of the Week'' because of its high popularity and rating on Steam. 6 years after its public release the mod has been played by more than 25 000 unique users and over 250 000 people have look at its ``mod page''. Finally, it has kept a consistent 5-Star Rating on Steam with 406 different reviews. 
In 2022 the mod was updated to keep up with the most recent versions of the original game, namely ``Skyrim Special Edition''. 

%Additionally there are more than 400 comments, from all platforms including Steam, the Nexus community and Reddit. We received comments from both mod users and more importantly, mod creators. The vast majority of which provided very encouraging feedback and suggestions, questions about implementation and authoring and a lot of requests about implementing it in other languages such as German, Russian and even Japanese. 

\section{Implementing CiF-CK in Unreal Engine 4 and Conan Exiles}
Conan Exiles \cite{conan} is an open-world survival video game published and developed by Funcom using a modified Unreal Engine 4 (UE4), the so-called DevKit. Conan Exiles was released as an early access title for the PC and Xbox One in 2017 and at the time of writing has full release planned in 2018 for both platforms plus the Playstation 4. In this section we will give some details on the ongoing implementation of CiF-CK in Conan Exiles. The DevKit has however one limitation: despite providing most of the tools that the developers use to make the game, it does not permit modders to interact with the game’s code. This means that some classes are inaccessible and can’t be changed directly.

In Conan Exiles by default most human NPCs are hostile and attack the player on sight. While it is possible to find a few non-aggressive NPCs in the game (including the ones the player captures),  the social interactions with them are very limited. They can be interacted with a button press, whoever their response is a random sentence from a pool of predefined general sentences, shared by every character, so no real dialogue is possible in the game’s current state. As it stands, Conan Exiles presents itself as a testbed with potential for social artificial intelligence testing and allows us to implement CiF-CK on the UE4 engine.

The assets in UE4 are available through Blueprints Visual Scripting interface, designed to define object-oriented classes and objects using a graphic node-based interface. Blueprints are a visual alternative to coding and can be used to define game events, new functions, classes and instancing them. By combining different nodes, blueprints are a powerful tool to add functionality to the game. Actors are objects in UE4 that can be placed in the world. They support geometric transformations like translate, rotate and scale as well as spawning and destroying operations, called using blueprints or C++ code. Components are instanced sub-objects within actors that define the actor’s behavior or functionality. Components can be swapped in order to change the actor’s behavior while the game is running. As components are instanced, components of the same type are independent from each other and any action performed on one of the instances is only reflected on that one. In comparison with Skyrim’s Creation Kit, Conan Exiles DevKit permits the use of more complex data structures and to control NPC’s lower level behaviors. On the other hand, dialogue and quest support are much weaker. These different characteristics change the way CiF-CK architecture is implemented in UE4.

The CiF-CK characters are an extension of the game’s Base Humanoid NPC class, which already supports gender. Traits are be implemented as a Map from string to boolean. Statuses are more complex structures mapping strings (e.g.: ``Angry\_at'') to an array of characters ID. Each social network are implemented as 3 maps (Value, Goal, Belief), each from known (by the NPC) characters to the respective values. Relationships are similarly represented, but without a belief, as their value is public knowledge.

Because the game offers limited dialogue support, a dialogue system must be implemented. It will be inspired by the one in \cite{oliveira2019model}. Regarding social exchanges, they will be implemented as components to be added to a character. Action towards another NPC will trigger events that update the social status, activate dialogue and apply the Trigger Rules. The CiF-CK script that handles decision making will be also  implemented using an UE4 component. This component will calculate volitions. The Initiator Influence Rules will be calculated in the initiator’s CiF-CK script component and the Responder Influence Rules in the responder’s component. The current implementation does not yet fully implement the cultural knowledge base and social facts database in the way that is proposed in the CiF architecture. 

%While this project is not finished yet, it shows that it is possible and relatively simple to implement the CiF-CK architecture in a general-purpose game engine.

\section{Conclusion}
Using a social simulation architecture, we have shown through the Social NPCs mod that it is possible to create rich and emergent NPC interactions. The benefit of using this architecture is that we are able to provide design patterns that can be elegantly implemented to represent character traits, statuses, and exchanges. With this architecture the authorial burden is shifted from code level to high level social interaction design which allows designers to focus on interaction design rather than AI implementation.

There are a number of ways in which the presented architecture can be improved and adapted. The vocabulary of traits, statuses, and behaviors can be extended. Authoring tools for specifying update rules and learning rule weights from example interactions is another exciting area of exploration. Finally, the complexity of the underlying model would be more visible and apparent to players if the dialog generation system could utilize the parameters of the model to generate more expressive dialogue.

\textbf{Author's Note:} It is important to note that CIF-CK has been successfully integrated in recent versions of FAtiMA-Toolkit \cite{mascarenhas2022fatima} as part of the author's work in its Doctoral Thesis \cite{guimaraes2019accessible}. Learn more about our work in the follow up paper (also released on Arxiv, for now) \cite{guimaraes2022towards}.

\section*{Acknowledgments}
This work has been partially supported by national funds through Funda\c c\~ao para a Ci\^encia e a Tecnologia (FCT) with reference UID/CEC /50021/2013 and project SLICE PTDC/CCI-COM/30787/2017 and by the EC H2020 project RAGE (Realising an Applied Gaming Eco-System), 
http://www.rageproject.eu/; Grant agreement No 644187.

\bibliographystyle{splncs04}
 \bibliography{mainbibliography}

\end{document}